\documentclass[conference]{IEEEtran}
\IEEEoverridecommandlockouts
\usepackage{cite}
\usepackage{amsmath,amssymb,amsfonts}
\usepackage{algorithmic}
\usepackage{graphicx}
\usepackage{textcomp}
\usepackage{xcolor}
\usepackage{amsmath,epsfig,amssymb,bbding}
\usepackage{cite}
\usepackage{multirow}
\usepackage{colortbl}
\usepackage[normalem]{ulem}
\useunder{\uline}{\ul}{}
\usepackage{booktabs}
\usepackage{doi}
\def\BibTeX{{\rm B\kern-.05em{\sc i\kern-.025em b}\kern-.08em
    T\kern-.1667em\lower.7ex\hbox{E}\kern-.125emX}}
\begin{document}

\title{Space-time Reinforcement Network for Video Object Segmentation

}

\author{\IEEEauthorblockN{1\textsuperscript{st} Yadang Chen}
\IEEEauthorblockA{\textit{School of Computer Science} \\
\textit{Nanjing University of Information Science and Technology}\\
Nanjing, China}
\and
\IEEEauthorblockN{2\textsuperscript{nd} Wentao Zhu}
\IEEEauthorblockA{\textit{School of Computer Science} \\
\textit{Nanjing University of Information Science and Technology}\\
Nanjing, China}
\and
\IEEEauthorblockN{\hspace{2em}3\textsuperscript{rd} Zhi-Xin Yang}
\IEEEauthorblockA{\hspace{2em}\textit{State Key Laboratory of Internet of Things for Smart City} \\
\hspace{2em}\textit{University of Macau}\\
  \hspace{2em}Macau, China}
\and
\IEEEauthorblockN{\hspace{9em}4\textsuperscript{th} Enhua Wu}
\IEEEauthorblockA{\hspace{9em}\textit{Key Laboratory of System Software and State Key Laboratory of Computer Science} \\
\textit{\hspace{9em}Institute of Software, Chinese Academy of Sciences}\\
\hspace{10em}Beijing, China}
}

\maketitle

\begin{abstract}
Recently, video object segmentation (VOS) networks typically use memory-based methods: for each query frame, the mask is predicted by space-time matching to memory frames. Despite these methods having superior performance, they suffer from two issues: 1) Challenging data can destroy the space-time coherence between adjacent video frames. 2) Pixel-level matching will lead to undesired mismatching caused by the noises or distractors. To address the aforementioned issues, we first propose to generate an auxiliary frame between adjacent frames, serving as an implicit short-temporal reference for the query one. Next, we learn a prototype for each video object and prototype-level matching can be implemented between the query and memory. The experiment demonstrated that our network outperforms the state-of-the-art method on the DAVIS 2017, achieving a $\mathcal{J}\&\mathcal{F}$ score of 86.4\%, and attains a competitive result 85.0\% on YouTube VOS 2018. In addition, our network exhibits a high inference speed of 32+ FPS.
\end{abstract}


\section{Introduction}
Semi-supervised video object segmentation (VOS) stands as a challenging task in computer vision, drawing widespread attention in autonomous driving, robotics and video editing. The key to this task is to fully utilize the given limited signals between frames, where the first-frame annotation is provided by the user, and it segments objects in the remaining frames as accurately as possible.

Recently, matching-based methods \cite{match1,chen3,match2} have achieved great success in semi-supervised VOS, achieving object segmentation through pixel-level matching between query frames and memory frames. Among these methods, memory-based networks \cite{1,2,3,qihe,4,5,6,7,chen2,9,chen1,10,mem1,mem2,mem3} attracted a lot of attention. For instance, the space-time storage network (STM) \cite{1} firstly is proposed to construct a feature memory for each object, and applies space-time matching between query and memory frames, which can better solve the problems of object occlusion and drift. After STM \cite{1}, there are many variants \cite{4,5,6,7,chen1,mem1,mem2,mem3} developed aiming to improve accuracy, reduce memory usage, and so on. Particularly, the latest related state-of-the-art methods, known as STCN \cite{9} and XMem \cite{10}, share multiple memory stores for all objects to capture different spatial-temporal contexts, have achieved prominent performance. 

\begin{figure}[t]
  \centerline{\includegraphics[scale=0.6]{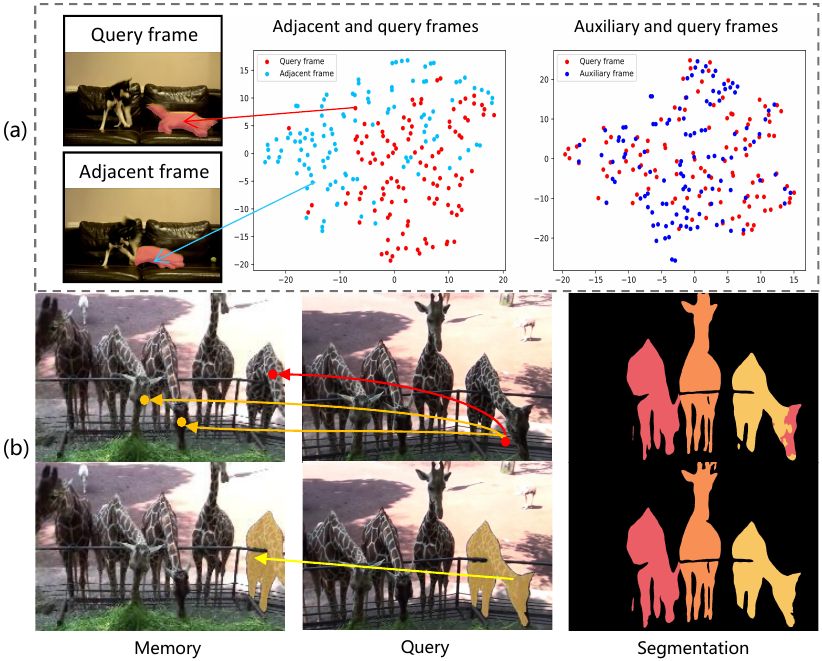}}
  \caption{(a) t-SNE visualization of the difference between frames. Left: the feature maps of query and adjacent frames. Right: the feature maps of query and auxiliary frames. Our proposed auxiliary frame is more consistent with the query than the adjacent frame. (b) Comparison of pixel-level matching (top) and prototype-level matching (bottom). Orange arrows indicate wrong matches. We propose prototype-level matching to improve undesired mismatching.}
  \label{fig:res1}
\end{figure}

Although these methods have achieved good results, there are still two issues that need to be carefully considered. Firstly, the existing methods \cite{3,7,10} investigate the space-time coherence between adjacent video frames to exclude distractor objects, however the coherence is often completely destroyed due to occlusion, fast motion and irregular deformation. This can be referred to Fig.\ref{fig:res1}(a), where the features between the query and its adjacent frame are inconsistent. Secondly, the dense pixel-level matching between the memory features and the query features will leads to undesired mismatching caused by the noises or distractors as shown in Fig.\ref{fig:res1}(b). Thus, a high-level matching is needful as an auxiliary manner for solving the problem.

Motivated by the above discussions, we propose a Space-time Reinforcement Network (SRNet) for video object segmentation to address the two mentioned weaknesses. In more details, the core ideas of this paper come from two aspects: i) Instead of directly exploring the space-time coherence between adjacent video frames, we propose to generate an auxiliary frame from adjacent frames, serving as an implicit short-temporal reference for the query one. The proposed auxiliary frame is more consistent with the query frame as shown in Fig.\ref{fig:res1}(a). ii) Beyond pixel-level matching, we propose to additionally learn a prototype for each video object. Thus, a high-level semantic matching, i.e., prototype-level matching, can be implemented between the query and memory to effectively improve the robustness of the method, also shown in Fig.\ref{fig:res1}(b).

Our contributions can be summarized as follows:

\begin{itemize}
  \item We propose to generate an implicit auxiliary frame between adjacent frames to improve the space-time coherence.
  \item We introduce a high-level semantic matching, i.e., prototype-level matching, to reduce undesired mismatching caused by the noises or distractors.
  \item Our SRNet achieves state-of-the-art performance on DAVIS 2017 and competitive results on YouTube 2018, while maintaining a high inference speed of 32+ FPS.

\end{itemize}

\begin{figure*}[t]
  \centerline{\includegraphics[scale=0.7]{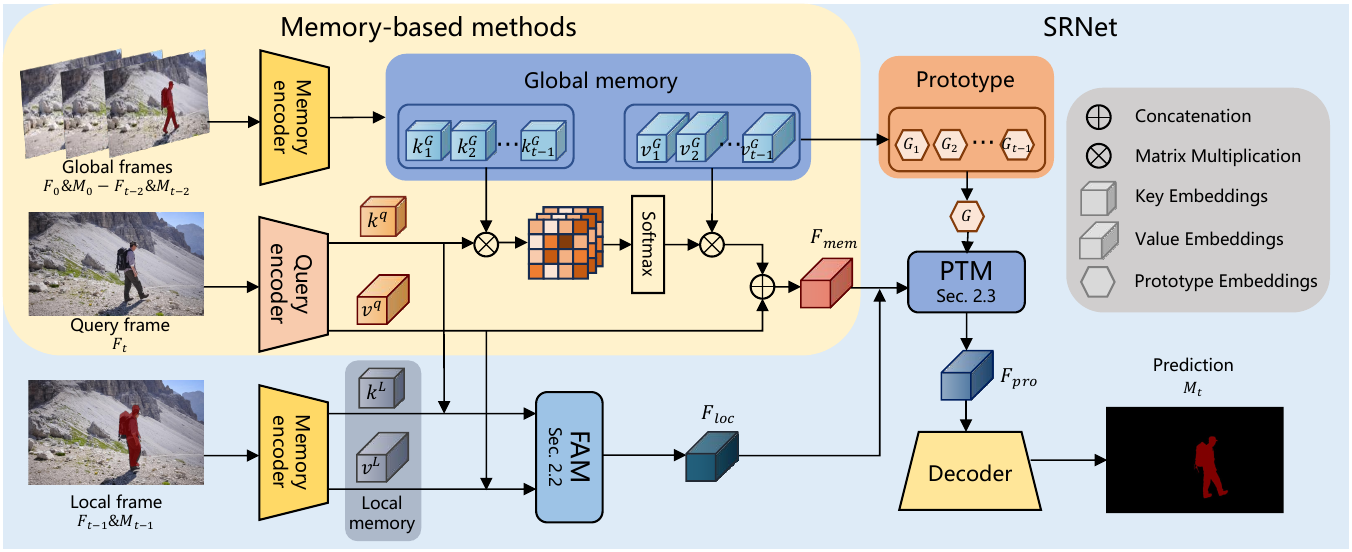}}
  \caption{An overview of SRNet. We propose a Feature Alignment Module (FAM) for generating an auxiliary frame to obtain the local feature and a Prototype Transformer Module (PTM) to implement prototype-level matching.}
\label{fig:res2}
\end{figure*}

\section{Method}
Fig.\ref{fig:res2} shows an overview of our proposed method SRNet. We firstly outline the recent memory-based advanced VOS methods. Furthermore, we elaborate on the improvements made by our SRNet based on these methods.

\noindent\textbf{Revisit memory-based VOS methods.} Since the seminal work of STM \cite{1}, memory-based methods have emerged as the predominant solution for VOS. Specifically, they use the current frame $F_{t} \in \mathbb{R}^{H^{0} \times W^{0} \times 3}$ as the query and store reference frames $\left\{ {F_{0},\cdots,F_{t-1}} \right\}$ and masks $\left\{ {M_{0},\cdots,M_{t - 1}} \right\}$ in memory, where $H^{0}$ and $W^{0}$ represent the initial size of the query frame. Then, the prediction of the mask $M^{t}$ can be obtained by space-time matching between query and memory. We revisit the framework of memory-based methods, e.g., STCN \cite{9} and Xmem \cite{10}, as they are one of the simplest and most effective memory-based methods.

As shown in Fig.2, given $T$ memory frames and a query frame, the query frame is passed to the query encoder to generate query key $k^{Q}\in\mathbb{R}^{H \times W \times C^{k}}$ and query value $v^{Q}\in\mathbb{R}^{H \times W \times C^{v}}$. Meanwhile, memory frames and masks are fed to the memory encoder to obtain memory key $k^{M} \in \mathbb{R}^{T \times H \times W \times C^{k}}$ and memory value $v^{M} \in \mathbb{R}^{T \times H \times W \times C^{v}}$, where $H$ and $W$ are spatial dimensions with a stride of 16, $C^{k}$ is the dimension of the key space and $C^{v}$ is the dimension of value space. Then, memory read performs space-time correspondence matching between $k^{Q}$ and $k^{M}$. We compute the affinity matrix $W \in \lbrack 0,1\rbrack^{THW \times HW}$:
\begin{eqnarray}
W\left( {i,j} \right) = \frac{exp\left( \varepsilon\left( {k^{M}(i),k^{q}(j)} \right) \right)}{\sum_{i}{exp\left( \varepsilon\left( {k^{M}(i),k^{q}(j)} \right) \right)}},
\end{eqnarray}
where $W\left( {i,j} \right)$ indicates the similarity between the i-th memory pixel and the j-th query pixel, and 
$\varepsilon\left( {\cdot,\cdot} \right)$ is a similarity measure, e.g., $L2$ distance.

Then, it multiplies $W\left( {i,j} \right)$ with $v^{M}$ to obtain the memory feature $F_{mem} \in \mathbb{R}^{H \times W \times C^{v}}$ for the query frame. Finally,  $F_{mem}$ is upsampled in decoder to generate prediction mask $M_{t} \in \mathbb{R}^{H^{0} \times W^{0} \times 1}$ .
\begin{figure}[t]
  \centerline{\includegraphics[scale=0.5]{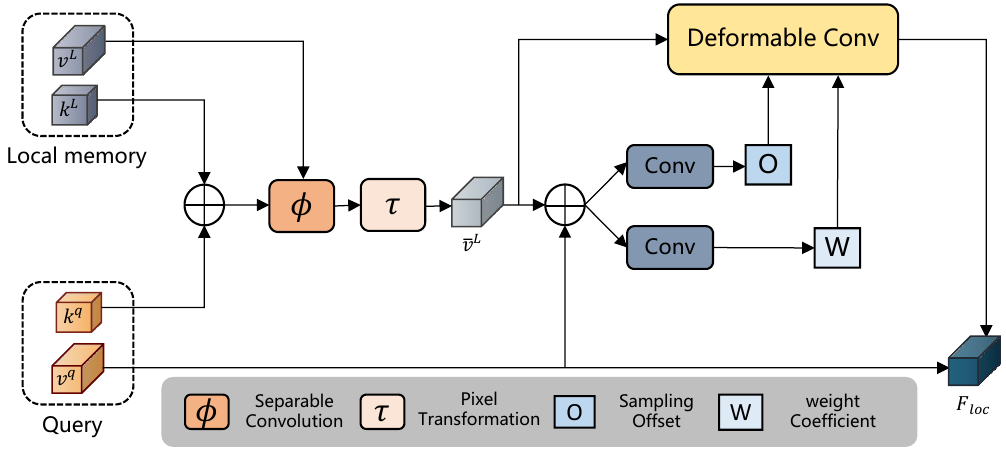}}
   \caption
  {Implementation of Feature Alignment Module.}
  \label{fig:res3}
\end{figure}

%\begingroup
%\setlength{\textfloatsep}{0.5em}

\subsection{Overview}\label{sec:srnet}
As shown in Fig.2, our SRNet divides the memory into two blocks, the local memory stores the adjacent frame $F_{t - 1}$ and its mask $M_{t-1}$, while the global memory holds frames $\left\{ {F_{0},\cdots,F_{t-2}} \right\}$ and their masks $\left\{ {M_{0},\cdots,M_{t - 2}} \right\}$ . Then we feed them  into the memory decoder to obtain local memory key $k^{L} \in \mathbb{R}^{H \times W \times C^{k}}$, local memory value $v^{L} \in \mathbb{R}^{H \times W \times C^{v}}$, global memory key $k^{G} \in \mathbb{R}^{(T-1)\times H \times W \times C^{k}}$ and global memory value $v^{G} \in \mathbb{R}^{(T-1)\times H \times W \times C^{v}}$.  We further learn memory prototype $G\in \mathbb{R}^{ 1 \times 1 \times C^{v}}$ by $v^{G}$.
Next, we pass $k^{L}$, $v^{L}$, $k^{Q}$ and $v^{Q}$ into a Feature Alignment Module (FAM) to obtain the value of auxiliary frame ${\overline{v}}^{L}$ and compute the local feature $F_{loc} \in \mathbb{R}^{H \times W \times C^{v}}$.
Then, the memory feature $F_{mem}$ is obtained by space-time matching beween query and global memory. Through a Prototype Transformer Module (PTM), pixel-level feature $F_{pix} \in \mathbb{R}^{H \times W \times C^{v}}$ , which is obtained by fusing $F_{loc}$ and $F_{mem}$ , are used for prototype-level matching with $G$. They are updated iteratively to obtain the final prototype feature $F_{pro} \in \mathbb{R}^{H \times W \times C^{v}}$. Finally, decoder upsamples $F_{pro}$ to generate prediction mask $M_{t}$.

\subsection{Feature Alignment Module}

We introduce FAM in this section, as shown in Fig.\ref{fig:res3}, which aligns the adjacent frame with the query frame to obtain an auxiliary frame, serving as an implicit short-temporal reference for the query one. We do not pursue image-level alignment, but rather learn feature-level alignment between the adjacent frame and query frame. Specifically, what we are learning is not the auxiliary frame, but its value ${\overline{v}}^{L}$. Subsequently, We refine ${\overline{v}}^{L}$ to obtain the local feature $F_{loc}$ of the query frame.

Given $k^{Q}$, $v^{Q}$, $k^{L}$ and  $v^{L}$, a uniform grid of points $ p \in \mathbb{R}^{H_{G} \times W_{G} \times 2}$ is generated as a reference. Specifically, the grid size is downsampled by a factor $g$ based on the input feature map size, $H_{G} = H/g$, $W_{G} = W/g$. The value of the reference point is a linearly spaced 2D coordinate $\left\{ (0,0), \cdot \cdot \cdot ,\left( H_{G} - 1,W_{G} - 1 \right) \right\}$, and utilizing the grid shape $H_{G}\times W_{G}$ is normalized to a range of [-1,+1], where (+1,+1) represents the lower right corner and (-1,-1) represents the upper left corner. In order to obtain the offset of each reference point, $k^{Q}$ and $k^{L}$ are input into the lightweight separable convolution $\phi~( \cdot )$ to generate the offset $\mathrm{\Delta}p$. Then, the local memory value $v^{L}$ and the offset $\mathrm{\Delta}p$ are inputted into the sampling function $\tau(\cdot,\cdot)$ to generate the value of auxiliary frame ${\overline{v}}^{L}$.

\begin{eqnarray}
\mathrm{\Delta}p = \phi\left( {k^{Q} \oplus k^{L}} \right),\overline{v}^{L} = \tau\left( v^{L},p + \mathrm{\Delta}p \right),
\end{eqnarray}
Specifically, we will set $\tau(\cdot,\cdot)$ to bilinear interpolation to make it differentiable:

\begin{eqnarray}
\tau\left({v^{L};\left( {p_{x},p_{y}} \right)} \right) = {\sum\limits_{({v_{x},v_{y}})}{g\left( {p_{x},v_{x}} \right)}}g\left( {p_{y},v_{y}} \right){v^{L}\left\lbrack v_{y},v_{x},: \right\rbrack},
\end{eqnarray}
where $g(a,~b)~ = ~max\left( 0,1 -  \middle| a - b \middle| \right)$ and $\left( {v_{x},v_{y}} \right)$ indices  all positions on  $v^{L} \in \mathbb{R}^{H \times W \times C^{v}}$.

Given ${\overline{v}}^{L}$ and $v^{Q}$, we estimate the convolutional kernel sampling offset $O$ and weight coefficient $W$ :
\begin{eqnarray}
O = conv\left( {\overline{v}}^{L}\oplus v^{Q} \right),\nonumber \\
W = conv\left( {\overline{v}}^{L}\oplus v^{Q} \right),
\end{eqnarray}
where $conv\left(\cdot\right)$ is a convolutional layer with a kernel size of 3 × 3, the sampling offset $O$ and weight coefficient $W$ respectively represent the positional shift and intensity fluctuation of each pixel in ${\overline{v}}^{L}$ relative to $v^{Q}$.
\begin{figure}[t]
  \centerline{\includegraphics[scale=0.5]{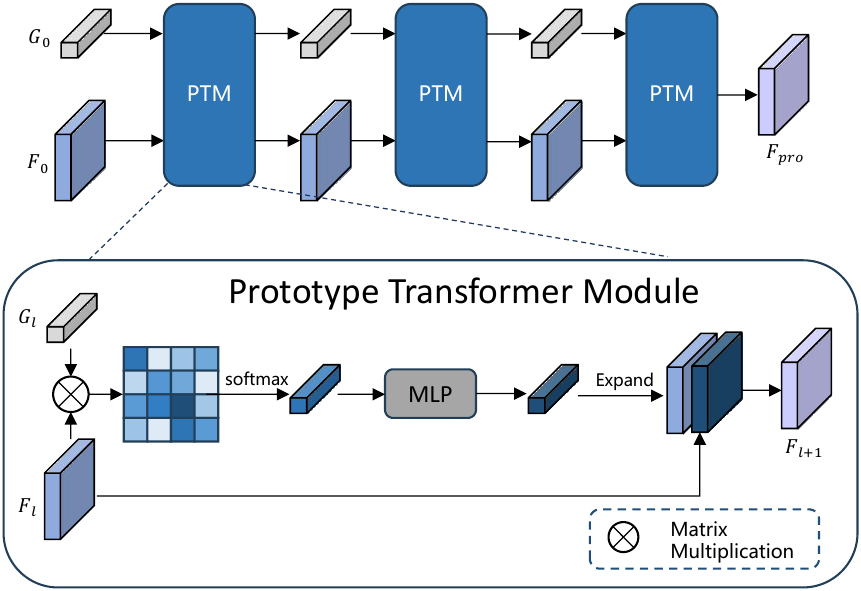}}
  \caption{Implementation of Prototype Transformer Module.}
  \label{fig:res4}
\end{figure}
Subsequently, given the value of the auxiliary frame ${\overline{v}}^{L}$, sampling offset $O$ and weight coefficient $W$ as inputs, local feature $F_{loc}$ can be computed:
\begin{eqnarray}
F_{loc}= deconv\left( {\overline{v}}^{L},O,W \right),
\end{eqnarray}
where $deconv\left(\cdot,\cdot,\cdot\right)$ is a modulated deformable convolution layer\cite{deformable}.

\setlength{\tabcolsep}{2.6pt}
\begin{table*}[]
\small %
\begin{center}
\caption{Quantitative comparison of YouTube 2018/2019 validation set, DAVIS 2017/2016 validation set, and 2017 test set. $S$ and $U$ represent visible or invisible categories. Red represents the best result, while blue represents the second-best result.} \label{tab:cap1}
\begin{tabular}{ccccccccccccccccccccc}

\hline
\rowcolor[HTML]{E7E6E6} 
\cellcolor[HTML]{E7E6E6}                        & \multicolumn{5}{c}{\cellcolor[HTML]{E7E6E6}YT-VOS 2018 Val}                                                                                                                                          & \multicolumn{5}{c}{\cellcolor[HTML]{E7E6E6}YouTube 2019 Val}                                                                                                                                           & \multicolumn{3}{c}{\cellcolor[HTML]{E7E6E6}DAVIS 2017 Val}                                                                     & \multicolumn{3}{c}{\cellcolor[HTML]{E7E6E6}DAVIS 2016 Val}                                                                     & \multicolumn{4}{c}{\cellcolor[HTML]{E7E6E6}DAVIS 2017 Test}                                                                             \\ 
\rowcolor[HTML]{E7E6E6} 

\multirow{-2}{*}{\cellcolor[HTML]{E7E6E6}Model} & $\mathcal{G}$                              & $\mathcal{J}_{s}$                                & $\mathcal{F}_{s}$                                 & $\mathcal{J}_{u}$                                 & $\mathcal{F}_{u}$        &$\mathcal{G} $ &$\mathcal{J}_{s}$                               & $\mathcal{F}_{s}$                                 & $\mathcal{J}_{u}$                     
& $\mathcal{F}_{u}$         & $\mathcal{J}\&\mathcal{F}$                                 & $\mathcal{J}$                                & $\mathcal{F}$         & $\mathcal{J}\&\mathcal{F}$                                 & $\mathcal{J}$                                 & $\mathcal{F}$         & $\mathcal{J}\&\mathcal{F}$                                 & $\mathcal{J}$                                & $\mathcal{J}\&\mathcal{F}$                                 & FPS                           \\ 
\hline
STM \cite{1}                                        & 79.4                            & 79.7                              & 84.2                              & 72.8                              & \multicolumn{1}{c|}{80.9}                              & -                                 & -                                 & -                                 & -                                 & \multicolumn{1}{c|}{-}                                 & 81.8                              & 79.2                              & \multicolumn{1}{c|}{84.3}                              & 89.3                              & 88.7                              & \multicolumn{1}{c|}{89.9}                              & 72.2                              & 69.3                              & 75.2                              & 11.1                        \\
CFBI \cite{2}                                           & 81.4                            & 81.1                              & 85.8                              & 75.3                              & \multicolumn{1}{c|}{83.4}                              & -                                 & -                                 & -                                 & -                                 & \multicolumn{1}{c|}{-}                                 & 81.9                              & 79.1                              & \multicolumn{1}{c|}{84.6}                              & 89.4                              & 88.3                              & \multicolumn{1}{c|}{90.5}                              & 78.0                               & 74.4                              & 81.6                              & 5.9                         \\
RMNet \cite{3}                                           & 81.5                            & 82.1                              & 85.7                              & 75.7                              & \multicolumn{1}{c|}{82.4}                              & -                                 & -                                 & -                                 & -                                 & \multicolumn{1}{c|}{-}                                 & 83.5                              & 81.0                              & \multicolumn{1}{c|}{86.0}                                & 88.8                              & 88.9                              & \multicolumn{1}{c|}{88.7}                              & 75.0                                & 71.9                              & 78.1                              & 4.4                         \\
HMMN \cite{4}                                           & 82.6                            & 82.1                              & 87.0                                & 76.8                              & \multicolumn{1}{c|}{84.6}                              & 82.5                              & 81.7                              & 86.1                              & 77.3                              & \multicolumn{1}{c|}{85.0}                                & 84.7                              & 81.9                              & \multicolumn{1}{c|}{87.5}                              & 90.8                              & 89.6                              & \multicolumn{1}{c|}{92.0}                                & 78.6                              & 74.7                              & 82.5                              & 9.3                         \\
STCN \cite{9}                                            & 83.0                              & 81.9                              & 86.5                              & 77.9                              & \multicolumn{1}{c|}{85.7}                              & 82.7                              & 81.1                              & 85.4                              & 78.2                              & \multicolumn{1}{c|}{85.9}                              & 85.4                              & 82.2                              & \multicolumn{1}{c|}{88.6}                              & {\color[HTML]{4472C4} {\ul 91.6}} & {\color[HTML]{FF0000} 90.8}       & \multicolumn{1}{c|}{{\color[HTML]{4472C4} {\ul 92.5}}} & 76.1                              & 73.1                             & 80.0                               & 20.2                        \\
AOT \cite{5}                                             & 84.1                            & 83.7                              & 88.5                              & 78.1                              & \multicolumn{1}{c|}{86.1}                              & 84.1                              & 83.5                              & 88.1                              & 78.4                              & \multicolumn{1}{c|}{86.3}                              & 84.9                              & 82.3                              & \multicolumn{1}{c|}{87.5}                              & 91.1                              & 90.1                              & \multicolumn{1}{c|}{92.1}                              & 78.8                              & 75.3                              & 82.3                              & 12.1                        \\
RDE \cite{6}                                            & -                               & -                                 & -                                 & -                                 & \multicolumn{1}{c|}{-}                                 & 81.9                              & 81.1                              & 85.5                              & 76.2                              & \multicolumn{1}{c|}{84.8}                              & 84.2                              & 80.8                              & \multicolumn{1}{c|}{87.5}                              & 91.1                              & 89.7                              & \multicolumn{1}{c|}{92.5}                              & 77.4                              & 73.6                              & 81.2                              & 27.0                          \\
XMem \cite{10}                                           & {\color[HTML]{FF0000} 85.7}     & {\color[HTML]{FF0000} 84.6}       & {\color[HTML]{FF0000} 89.3}       & {\color[HTML]{FF0000} 80..2}      & \multicolumn{1}{c|}{{\color[HTML]{FF0000} 88.7}}       & {\color[HTML]{FF0000} 85.5}       & {\color[HTML]{FF0000} 84.3}       & {\color[HTML]{FF0000} 88.6}       & {\color[HTML]{FF0000} 80.3}       & \multicolumn{1}{c|}{{\color[HTML]{FF0000} 88.6}}       & {\color[HTML]{4472C4} {\ul 86.2}} & {\color[HTML]{4472C4} {\ul 82.9}} & \multicolumn{1}{c|}{{\color[HTML]{4472C4} {\ul 89.5}}} & 91.5                              & 90.4                              & \multicolumn{1}{c|}{92.7}                              & {\color[HTML]{FF0000} {81.0}} & {\color[HTML]{FF0000} {77.4}} & {\color[HTML]{FF0000} {84.5}} & 22.6                        \\ \hline
SRNet                                           & {\color[HTML]{4472C4} {\ul 85.0}} & {\color[HTML]{4472C4} {\ul 83.9}} & {\color[HTML]{4472C4} {\ul 88.5}} & {\color[HTML]{4472C4} {\ul 79.7}} & \multicolumn{1}{c|}{{\color[HTML]{4472C4} {\ul 87.9}}} & {\color[HTML]{4472C4} {\ul 84.7}} & {\color[HTML]{4472C4} {\ul 82.9}} & {\color[HTML]{4472C4} {\ul 87.4}} & {\color[HTML]{4472C4} {\ul 80.3}} & \multicolumn{1}{c|}{{\color[HTML]{4472C4} {\ul 88.3}}} & {\color[HTML]{FF0000} 86.4}       & {\color[HTML]{FF0000} 83.1}       & \multicolumn{1}{c|}{{\color[HTML]{FF0000} 89.7}}       & {\color[HTML]{FF0000} 91.8}       & {\color[HTML]{4472C4} {\ul 90.6}} & \multicolumn{1}{c|}{{\color[HTML]{FF0000} 93.0}}        &{\color[HTML]{4472C4} {\ul 79.2}}                             &{\color[HTML]{4472C4} {\ul 76.5}}                              &{\color[HTML]{4472C4} {\ul 81.9}}                                & {\color[HTML]{FF0000} 32.3} \\ \hline
\end{tabular}
\end{center}
\end{table*}

\begin{figure*}[t]
  \centerline{\includegraphics[scale=0.65]{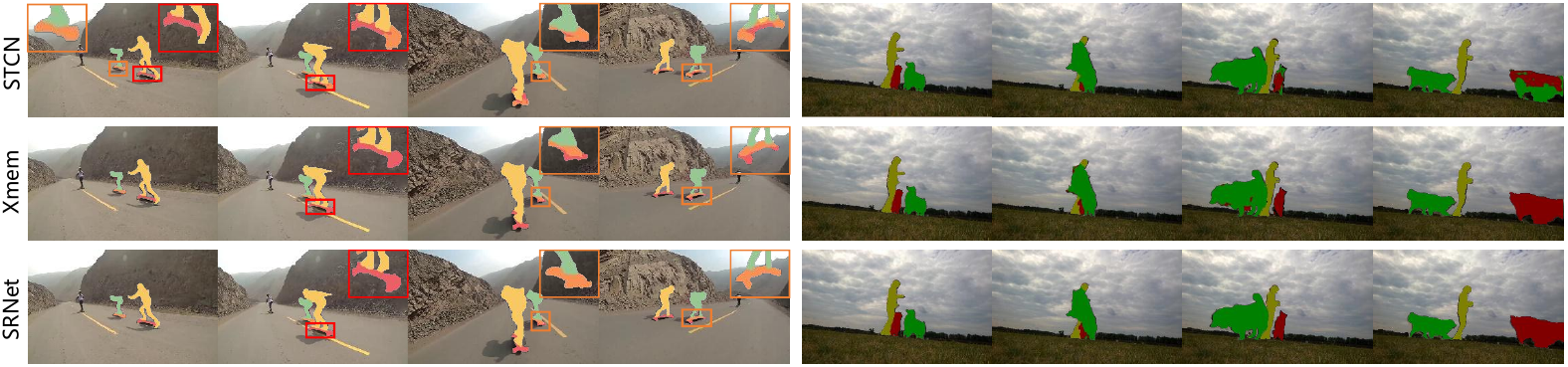}}
  \caption{Qualitative comparisons with SRNet, STCN \cite{9} and Xmem \cite{10} on the YouTube 2018 validation set and DAVIS 2017 validation set.}
  \label{fig:res5}
\end{figure*}

\setlength{\tabcolsep}{6pt}
\begin{table}[t]
\begin{center}
\caption{Module ablation study.} \label{tab:cap3}
\begin{tabular}{ccccc}
\rowcolor[HTML]{E7E6E6} 
\hline    
 FAM      &PTM      &$\mathcal{J}\&\mathcal{F}$   &$\mathcal{G}$ &FPS \\
\hline
\Checkmark   &\multicolumn{1}{c|}{\Checkmark} &86.4  &85.0 &32.3 \\
\hline
 &\multicolumn{1}{c|}{\Checkmark} &86.0 &84.6 &37.5 \\
\Checkmark &\multicolumn{1}{c|}{} &85.8 &84.3  &38.7 \\
\hline
\end{tabular}
\end{center}
\end{table}

\subsection{Prototype Transformer Module}

In fact, pixel-level matching will leads to undesired mismatching caused by the
noises or distractors. To address this issue, we propose a Prototype Transformer Module (PTM) as shown in Fig.\ref{fig:res4} which performs prototype-level matching between query and memory.

We can use mask annotations on the memory frames to learn the foreground prototype. There are two strategies for utilizing segmentation masks, namely early fusion and late fusion. Early fusion will mask the supporting images before inputting them into the feature extractor. Late fusion directly masks feature maps to generate
foreground features separately. In this work, we adopt an early fusion strategy because we can reuse the $v^{G}$ obtained through the memory encoder.

Specifically, given $v^{G}$, the formula for calculating the initial prototype $G_{0}$ and initial pixel feature $F_{0}$:

\begin{eqnarray}
G_{0} = \frac{\sum_{i = 1}^{THW}{F\left( v_{i}^{G} \right)W\left( v_{i}^{G} \right)}}{\sum_{i = 1}^{THW}{W\left( v_{i}^{G} \right)}},
\end{eqnarray}

\vspace{-0.5cm}
\begin{eqnarray}
F_{0}= fusion\left( F_{mem},F_{loc} \right),
\end{eqnarray}
where, $F(\cdot)$ is a 2-layer, $C^{v}$-dimensional MLP,  $W(\cdot)$ is a 2-layer, 1-dimensional MLP and $fusion\left( \cdot,\cdot\right)$  includes two ResBlocks and a CBAM block.

Our PTM has learnable prototype $G$ and pixel feature $F$, which iteratively updated by initial prototype $G_{0}$ and initial pixel feature $F_{0}$  through cross attention. We added position embeddings to the key of each attention layer.
\begin{eqnarray}
q = G_{0}W_{q},~k~ = F_{0}W_{k},v~ = F_{0}W_{v},
\end{eqnarray}

\vspace{-0.7cm}

\begin{eqnarray}
G_{1} = MLP\left( softmax\left( q\left(k + P_{k} \right)^{T} \right)v\odot G_{0} \right),
\end{eqnarray}
where, $W_{q}$ , $W_{k}$ , $W_{v}$ , $P_{k}$ is the learnable parameters in SPA.

Then, guided by the updated prototype $G$, the semantic information of the corresponding pixels in pixel feature $F$ is activated. Specifically, $G$ is extended and combined with $F$  to activate the target area:
\begin{eqnarray}
F_{1} = \varphi\left( expand\left(G_{0}\right)\odot F_{0} \right),
\end{eqnarray}
where $\varphi(\cdot)$ is a simple activation network consisting of two 3 × 3 convolutional layers and one channel attention layer.

Finally, we use $G_{1}$ as the new $G$ and $F_{1}$ as the new $F$, and update them three times to obtain the final prototype feature $F_{pro}$.

\begin{figure}[t]
  \centerline{\includegraphics[scale=0.6]{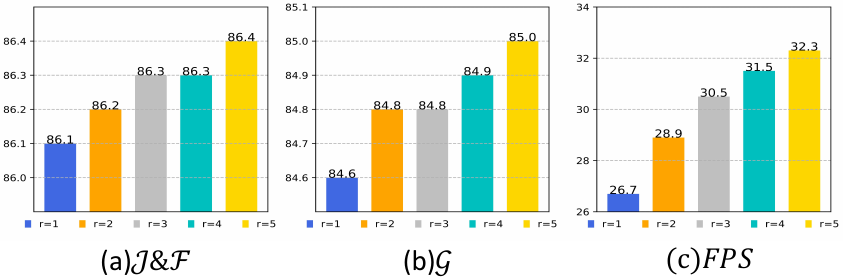}}
  \caption{Ablation study of different sampling intervals on local memory in FAM.}
  \label{fig:res6}
\end{figure}

\section{Experiment}
For assessment purposes, we utilize standard metrics: the Jaccard index $\mathcal{J}$, contour accuracy $\mathcal{F}$, and their combined average $\mathcal{J}\&\mathcal{F}$. In YouTubeVOS \cite{11}, the computation of $\mathcal{J}$ and $\mathcal{F}$ is performed separately for ``seen" and ``unseen" categories. $\mathcal{G}$ represents the averaged $\mathcal{J}\&\mathcal{F}$ across both ``seen" and ``unseen" categories.

\subsection{Quantitative Comparison}
Table \ref{tab:cap1} tabulate our results on YouTube 2018/2019 \cite{11} validation, DAVIS 2016/2017 \cite{12} validation, and DAVIS 2017 test-dev. On the DAVIS 2017 and 2016 validation set, our SRNet outperforms the baseline STCN \cite{9} by 1\% and 0.2\% in $\mathcal{J}\&\mathcal{F}$ and runs about 60\% faster and outperforms the state-of-the-art method Xmem \cite{10} by 0.2\% and 0.3\% in $\mathcal{J}\&\mathcal{F}$ and runs about 43\% faster. On the YouTube 2018 validation and DAVIS 2017 test-dev set, SRNet performs 2\% and 2.1\% better on $\mathcal{J}\&\mathcal{F}$ than STCN \cite{9}, but 0.7\% and 1.8\% less than Xmem \cite{10}. Thanks to setting our $C^{v}$ at 256, rather than 512, our inference speed far exceeds other methods. We choose STCN \cite{9} as the baseline because its GPU usage is 20\% lower than XMem \cite{10} and nearly twice as fast when training.

 \begin{figure}[t]
  \centerline{\includegraphics[scale=0.4]{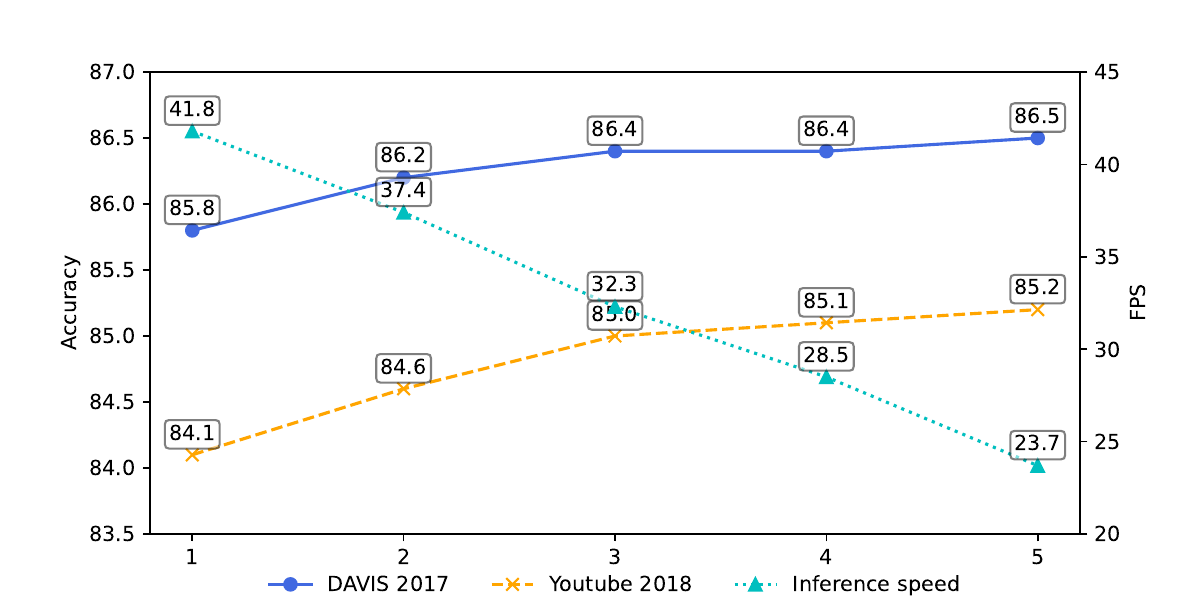}}
  \caption{Ablation study of different numbers of PTM layers.}
  \label{fig:7}
\end{figure}

\begin{figure}[t]
  \centerline{\includegraphics[scale=0.65]{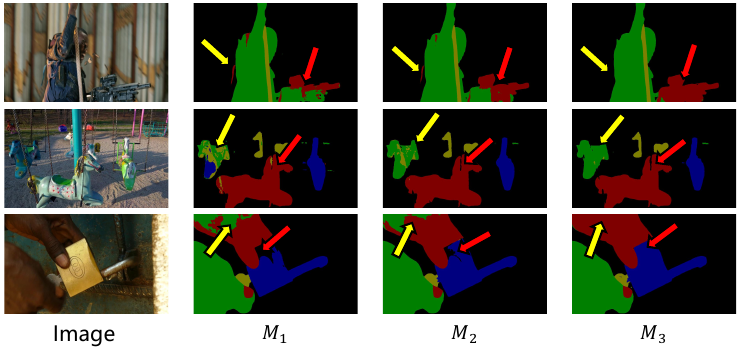}}
  \caption{Visualization of auxiliary masks at different layers of PTM.}
  \label{fig:8}
\end{figure}

\subsection{Qualitative Comparison}
 We choose two videos as examples, one from DAVIS 2017 validation and another from YouTube 2018. and present the segmentation results in comparison with SRNet, STCN \cite{9} and Xmem \cite{10} in Fig.\ref{fig:res5}. As can be seen, our SRNet produces more accurate masks: similar objects can be distinguished, and undesired mismatching can be reduced.

\subsection{Ablations}
We conduct ablation experiments on DAVIS 2017 validation and YouTube 2018 validation. In particular, we report $\mathcal{J}\&\mathcal{F}$ and FPS for DAVIS 2017, alongside $\mathcal{G}$ for YouTube 2018. 

Ablation study on the two modules of SRNet is conducted to evaluate their individual effectiveness, as shown in Table \ref{tab:cap3}. When two modules are activated (i.e., in the default configuration), scoring 86.4\% on DAVIS 2017 and 85.0\% on YouTube 2018, with an inference speed reaching 32.3 FPS. When FAM is disabled, the result decreases 0.4\% and 0.4\%. Then, When PTM is disabled, the result decreases 0.6\% and 0.7\%.

To verify the impact of sampling intervals of local memory in FAM, we choose different intervals for the experiment, as shown in the Fig.\ref{fig:res6}.

\begin{figure}[t]
  \centerline{\includegraphics[scale=0.65]{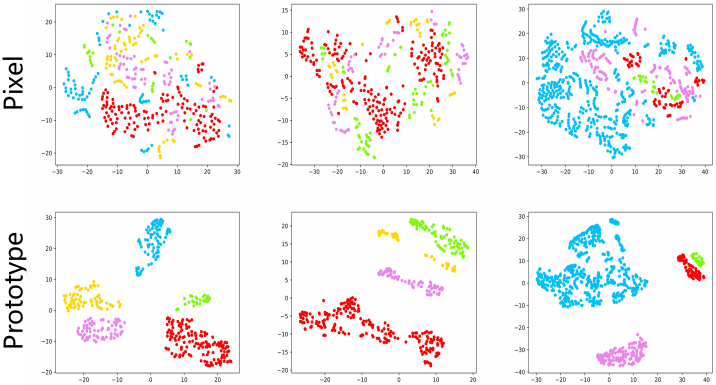}}
  \caption{Visualization results to compare prototype feature (bottom) with pixel feature (top) on three videos (three columns). Varying colors correspond to foreground pixels from different classes. Each point represents a feature vector with shape of 1×1×$C^{v}$ at a position in the feature, where $C^{v}$ indicates the dimension of the value space.}
  \label{fig:9}
\end{figure}

An analysis of different numbers of PTM layers on the effectiveness of our network in Fig.\ref{fig:7}. In order to better balance accuracy and FPS, we choose to iterate three times. Additionally, we visualized each layer's auxiliary mask in Fig.\ref{fig:8}. It can be observed that the object becomes more coherent (red arrows), and errors are suppressed (yellow arrows).
In Fig.\ref{fig:9}, it can be seen that the distance between pixels of the same class in pixel feature is far, while the distance between pixels of different classes is close. Obviously,the prototype feature after PTM has effectively reduced intra-class variation and enlarged inter-class difference.

\begin{figure}[t]
  \centerline{\includegraphics[scale=0.5]{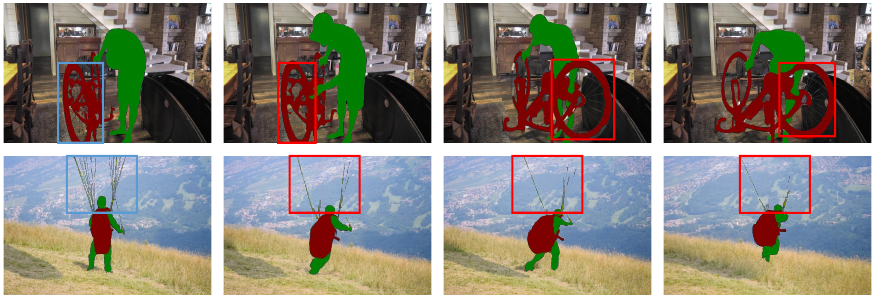}}
  \caption{Failure cases: the fine-grained objects are not well segmented. The first column is the first frame with ground truth. Error predictions are highlighted with red bounding boxes.}
  \label{fig:10}
\end{figure}

\subsection{Limitations}
 As shown in Fig.\ref{fig:10}, SRNet typically fails in small-scale foreground objects. We believe this is mainly because our method cannot capture fine-grained information. We hope to find more effective methods to distinguish small objects and achieve better performance in future work.

 \section{Conclusion}
In this paper, we tackle the issues of memory-based methods for semi-supervised video object segmentation (VOS). For improving  the space-time coherence, we first propose to generate an implicit auxiliary frame between adjacent frames. Secondly, we introduce a high-level semantic matching, i.e., prototype-level matching to reduce undesired mismatching caused by the noises or distractors. The experiment results show that our SRNet achieves competitive performance.

\section{Acknowledgement}
This work was supported in part by the National Natural Science Foundation of China under Grant 62332015, and Grant 62072449; in part by the Science and Technology Development Fund, Macau, SAR, under Grant 0075/2023/AMJ, Grant 0003/2023/RIB1, and Grant 001/2024/SKL; in part by the Zhuhai Science and Technology Innovation Bureau under Grant ZH2220004002524; and in part by the University of Macau under Grant MYRG2022-00059-FST and Grant MYRG-GRG2023-00237-FST-UMDF.

\bibliographystyle{IEEEtran}
\bibliography{icme2023template}

\begin{thebibliography}{10}
\providecommand{\url}[1]{#1}
\csname url@samestyle\endcsname
\providecommand{\newblock}{\relax}
\providecommand{\bibinfo}[2]{#2}
\providecommand{\BIBentrySTDinterwordspacing}{\spaceskip=0pt\relax}
\providecommand{\BIBentryALTinterwordstretchfactor}{4}
\providecommand{\BIBentryALTinterwordspacing}{\spaceskip=\fontdimen2\font plus
\BIBentryALTinterwordstretchfactor\fontdimen3\font minus \fontdimen4\font\relax}
\providecommand{\BIBforeignlanguage}[2]{{%
\expandafter\ifx\csname l@#1\endcsname\relax
\typeout{** WARNING: IEEEtran.bst: No hyphenation pattern has been}%
\typeout{** loaded for the language `#1'. Using the pattern for}%
\typeout{** the default language instead.}%
\else
\language=\csname l@#1\endcsname
\fi
#2}}
\providecommand{\BIBdecl}{\relax}
\BIBdecl

\bibitem{match1}
P.~Voigtlaender, Y.~Chai, F.~Schroff, H.~Adam, B.~Leibe, and L.-C. Chen, ``Feelvos: Fast end-to-end embedding learning for video object segmentation,'' in \emph{Proceedings of the IEEE/CVF Conference on Computer Vision and Pattern Recognition}, 2019, pp. 9481--9490.

\bibitem{chen3}
L.~Fu, Z.~Li, Q.~Ye, H.~Yin, Q.~Liu, X.~Chen, X.~Fan, W.~Yang, and G.~Yang, ``Learning robust discriminant subspace based on joint $\mathrm{L}_{2,p}$- and $\mathrm{L}_{2,s}$-norm distance metrics,'' \emph{IEEE transactions on neural networks and learning systems}, vol.~33, no.~1, pp. 130--144, 2020.

\bibitem{match2}
Z.~Yang, Y.~Wei, and Y.~Yang, ``Collaborative video object segmentation by foreground-background integration,'' in \emph{European Conference on Computer Vision}.\hskip 1em plus 0.5em minus 0.4em\relax Springer, 2020, pp. 332--348.

\bibitem{1}
S.~W. Oh, J.-Y. Lee, N.~Xu, and S.~J. Kim, ``Video object segmentation using space-time memory networks,'' in \emph{Proceedings of the IEEE/CVF International Conference on Computer Vision}, 2019, pp. 9226--9235.

\bibitem{2}
Z.~Yang, Y.~Wei, and Y.~Yang, ``Collaborative video object segmentation by multi-scale foreground-background integration,'' \emph{IEEE Transactions on Pattern Analysis and Machine Intelligence}, vol.~44, no.~9, pp. 4701--4712, 2021.

\bibitem{3}
H.~Xie, H.~Yao, S.~Zhou, S.~Zhang, and W.~Sun, ``Efficient regional memory network for video object segmentation,'' in \emph{Proceedings of the IEEE/CVF Conference on Computer Vision and Pattern Recognition}, 2021, pp. 1286--1295.

\bibitem{qihe}
Q.~Huang, L.~Shen, R.~Zhang, J.~Cheng, S.~Ding, Z.~Zhou, and Y.~Wang, ``Hdmixer: Hierarchical dependency with extendable patch for multivariate time series forecasting,'' in \emph{Proceedings of the AAAI Conference on Artificial Intelligence}, vol.~38, no.~11, 2024, pp. 12\,608--12\,616.

\bibitem{4}
H.~Seong, S.~W. Oh, J.-Y. Lee, S.~Lee, S.~Lee, and E.~Kim, ``Hierarchical memory matching network for video object segmentation,'' in \emph{Proceedings of the IEEE/CVF International Conference on Computer Vision}, 2021, pp. 12\,889--12\,898.

\bibitem{5}
Z.~Yang, Y.~Wei, and Y.~Yang, ``Associating objects with transformers for video object segmentation,'' \emph{Advances in Neural Information Processing Systems}, vol.~34, pp. 2491--2502, 2021.

\bibitem{6}
M.~Li, L.~Hu, Z.~Xiong, B.~Zhang, P.~Pan, and D.~Liu, ``Recurrent dynamic embedding for video object segmentation,'' in \emph{Proceedings of the IEEE/CVF Conference on Computer Vision and Pattern Recognition}, 2022, pp. 1332--1341.

\bibitem{7}
Y.~Chen, D.~Zhang, Z.-x. Yang, and E.~Wu, ``Robust and efficient memory network for video object segmentation,'' \emph{arXiv preprint arXiv:2304.11840}, 2023.

\bibitem{chen2}
Q.~Ye, P.~Huang, Z.~Zhang, Y.~Zheng, L.~Fu, and W.~Yang, ``Multiview learning with robust double-sided twin svm,'' \emph{IEEE transactions on Cybernetics}, vol.~52, no.~12, pp. 12\,745--12\,758, 2021.

\bibitem{9}
H.~K. Cheng, Y.-W. Tai, and C.-K. Tang, ``Rethinking space-time networks with improved memory coverage for efficient video object segmentation,'' \emph{Advances in Neural Information Processing Systems}, vol.~34, pp. 11\,781--11\,794, 2021.

\bibitem{chen1}
Y.~Chen, C.~Hao, Z.-X. Yang, and E.~Wu, ``Fast target-aware learning for few-shot video object segmentation,'' \emph{Science China Information Sciences}, vol.~65, no.~8, p. 182104, 2022.

\bibitem{10}
H.~K. Cheng and A.~G. Schwing, ``Xmem: Long-term video object segmentation with an atkinson-shiffrin memory model,'' in \emph{European Conference on Computer Vision}.\hskip 1em plus 0.5em minus 0.4em\relax Springer, 2022, pp. 640--658.

\bibitem{mem1}
M.~Li, L.~Hu, Z.~Xiong, B.~Zhang, P.~Pan, and D.~Liu, ``Recurrent dynamic embedding for video object segmentation,'' in \emph{Proceedings of the IEEE/CVF Conference on Computer Vision and Pattern Recognition}, 2022, pp. 1332--1341.

\bibitem{mem2}
K.~Park, S.~Woo, S.~W. Oh, I.~S. Kweon, and J.-Y. Lee, ``Per-clip video object segmentation,'' in \emph{Proceedings of the IEEE/CVF Conference on Computer Vision and Pattern Recognition}, 2022, pp. 1352--1361.

\bibitem{mem3}
Y.~Chen, D.~Zhang, Y.~Zheng, Z.-X. Yang, E.~Wu, and H.~Zhao, ``Boosting video object segmentation via robust and efficient memory network,'' \emph{IEEE Transactions on Circuits and Systems for Video Technology}, 2023.

\bibitem{deformable}
X.~Zhu, H.~Hu, S.~Lin, and J.~Dai, ``Deformable convnets v2: More deformable, better results,'' in \emph{Proceedings of the IEEE/CVF conference on computer vision and pattern recognition}, 2019, pp. 9308--9316.

\bibitem{11}
N.~Xu, L.~Yang, Y.~Fan, D.~Yue, Y.~Liang, J.~Yang, and T.~Huang, ``Youtube-vos: A large-scale video object segmentation benchmark,'' \emph{arXiv preprint arXiv:1809.03327}, 2018.

\bibitem{12}
J.~Pont-Tuset, F.~Perazzi, S.~Caelles, P.~Arbel{\'a}ez, A.~Sorkine-Hornung, and L.~Van~Gool, ``The 2017 davis challenge on video object segmentation,'' \emph{arXiv preprint arXiv:1704.00675}, 2017.

\end{thebibliography}

\end{document}